\documentclass{article}

\PassOptionsToPackage{numbers,sort&compress}{natbib}
\usepackage[preprint]{neurips_2026}
\makeatletter
\renewcommand{\@noticestring}{Preprint. Extended version of a paper accepted at the NeurIPS 2026 Workshop on Robot Learning
  with World Models: Capabilities, Frontiers, and Challenges.}
\makeatother

\usepackage[utf8]{inputenc}
\usepackage[T1]{fontenc}
\usepackage{float}
\usepackage{hyperref}
\usepackage{url}
\usepackage{booktabs}
\usepackage{amsmath}
\usepackage{amsfonts}
\usepackage{amssymb}
\usepackage{nicefrac}
\usepackage{microtype}
\usepackage{xcolor}
\usepackage{xspace}
\usepackage{graphicx}
\usepackage{wrapfig}
\usepackage{tikz}
\usetikzlibrary{arrows.meta,calc,decorations.pathreplacing,positioning}

\definecolor{cImg}{HTML}{CFDFF3}
\definecolor{cFut}{HTML}{E2D5F2}
\definecolor{cVec}{HTML}{F7E1C4}
\definecolor{cAct}{HTML}{EFA9A5}
\definecolor{cKeep}{HTML}{3B5B92}

\newcommand{\smol}{SmolWAM\xspace}
\newcommand{\smolla}{SmolWAM-LA\xspace}
\newcommand{\cosmos}{Cosmos Policy\xspace}

\title{SmolWAM: A Latent World-Action Model for\\Robotic Manipulation on Edge Devices}

\author{%
  Timi Owolabi \quad Hikmah Olawore \quad Frances Adelakun \quad Ayotomiwa Oyewumi\\
  ML Collective\\
  \texttt{\{timiiowolabi, olaworehikmah, francesadelakun127, ayot.oyewumi\}@gmail.com}
}

\begin{document}

\maketitle

\begin{abstract}
World-action models (WAMs) predict a robot's actions together with the observations those actions will produce. However, the strongest models rely on video diffusion transformers that are far too computationally expensive for resource-constrained edge devices.
We present \smol, a latent WAM that we derive from the 2B-parameter \cosmos by keeping its pretrained trunk and rebuilding its inference interface: single-frame visual latents, cached text features that preserve the trunk's attention sinks, 12 of 28 transformer blocks, and 1,008 tokens from which actions and future visual latents are read out in one rectified-flow step.
Trained in 3.6 GPU-hours, \smol succeeds in 95.5\% of 400 LIBERO episodes at 57\,ms and 2.3\,TFLOP per 16-action chunk with 2.4\,GiB of GPU memory, 11$\times$ faster than its teacher with 18.5$\times$ fewer FLOPs and 3.2$\times$ less memory, and its world head still forecasts scene motion.
Code: \href{https://github.com/iitimii/SmolWAM}{\texttt{github.com/iitimii/SmolWAM}}; project page: \href{https://smolwam.vercel.app/}{\texttt{smolwam.vercel.app}}.
\end{abstract}

%===============================================================================
\section{Introduction}
\label{sec:intro}

World-action models (WAMs) predict a robot's next actions together with the observations they should produce~\citep{kim2026cosmospolicy,ye2026dreamzero,lu2026wamsurvey}.
\cosmos~\citep{kim2026cosmospolicy} builds one from the Cosmos-Predict2 video model~\citep{nvidia2025cosmos} by treating actions, future observations, and values as latent frames of a single denoising sequence, and reaches 98.5\% success on LIBERO~\citep{liu2023libero}.
Its inference stack, however, was built for video generation: a 4.8B-parameter T5-XXL text encoder~\citep{raffel2020t5}, a VAE that encodes a 33-frame clip per query, and five denoising steps of a 2B-parameter transformer over 1,764 tokens, which cost 624\,ms, 42.6\,TFLOP, and 7.55\,GiB per 16-action chunk on an A100 even with cached text features.

Efficient policies usually give up the world model or the video prior: compact vision-language-action (VLA) models such as SmolVLA~\citep{shukor2025smolvla} and TurboVLA~\citep{xie2026turbovla} predict only actions, and Fast-WAM~\citep{yuan2026fastwam} skips future prediction at test time.
We ask whether a strong pretrained WAM can itself be made small while still predicting the future.

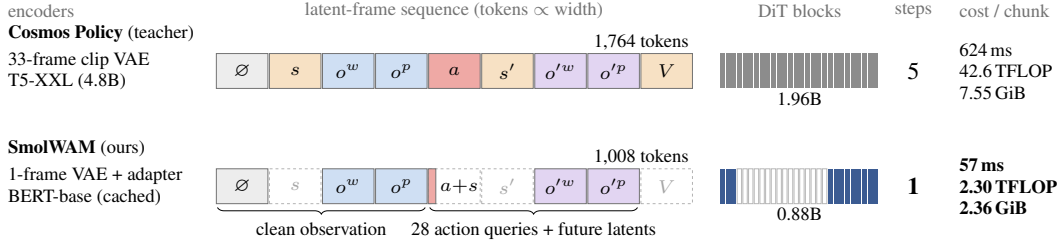
\begin{figure}[t]
\centering
\resizebox{\textwidth}{!}{% Figure 1: deriving SmolWAM from Cosmos Policy (interfaces, latent-frame layout, depth, steps, cost).
\begin{tikzpicture}[
  x=1cm, y=1cm, font=\scriptsize,
  slot/.style={draw=black!55, line width=0.3pt, minimum height=0.46cm, minimum width=0.68cm,
               inner sep=0pt, anchor=west},
  sblank/.style={slot, fill=black!7},
  svec/.style={slot, fill=cVec},
  simg/.style={slot, fill=cImg},
  sfut/.style={slot, fill=cFut},
  sact/.style={slot, fill=cAct},
  ghost/.style={slot, draw=black!30, dash pattern=on 1.1pt off 1.1pt, fill=white, text=black!35},
  brace/.style={decorate, decoration={brace, amplitude=2.5pt, mirror}, line width=0.35pt},
  inp/.style={anchor=west, align=left, inner sep=0pt},
  hdr/.style={text=black!55, inner sep=0pt},
]
\def\x#1{2.75+0.70*#1}   % left edge of latent-frame slot #1 (0..8)
\def\d#1{9.40+0.075*#1}  % left edge of DiT block #1 (0..27)

% column headers
\node[hdr, anchor=west] at (0,3.02) {encoders};
\node[hdr] at ({\x4+0.34},3.02) {latent-frame sequence (tokens $\propto$ width)};
\node[hdr] at (10.45,3.02) {DiT blocks};
\node[hdr] at (11.95,3.02) {steps};
\node[hdr, anchor=west] at (12.55,3.02) {cost / chunk};

% ---------------- Row 1: Cosmos Policy ----------------
\node[inp] at (0,2.72) {\textbf{Cosmos Policy} (teacher)};
\node[inp] at (0,2.22) {33-frame clip VAE\\ T5-XXL (4.8B)};
\node[sblank] at ({\x0},2.22) {$\varnothing$};
\node[svec]   at ({\x1},2.22) {$s$};
\node[simg]   at ({\x2},2.22) {$o^{w}$};
\node[simg]   at ({\x3},2.22) {$o^{p}$};
\node[sact]   at ({\x4},2.22) {$a$};
\node[svec]   at ({\x5},2.22) {$s'$};
\node[sfut]   at ({\x6},2.22) {$o'^{w}$};
\node[sfut]   at ({\x7},2.22) {$o'^{p}$};
\node[svec]   at ({\x8},2.22) {$V$};
\node[anchor=south east, inner sep=1pt] at ({\x8+0.68},2.46) {1,764 tokens};
\foreach \b in {0,...,27} { \fill[black!45] ({\d\b},1.99) rectangle ++(0.058,0.46); }
\node[anchor=north, inner sep=1pt] at (10.45,1.98) {1.96B};
\node at (11.95,2.22) {\small 5};
\node[inp] at (12.55,2.22) {624\,ms\\ 42.6\,TFLOP\\ 7.55\,GiB};

% ---------------- Row 2: SmolWAM ----------------
\node[inp] at (0,1.20) {\textbf{SmolWAM} (ours)};
\node[inp] at (0,0.70) {1-frame VAE + adapter\\ BERT-base (cached)};
\node[sblank] at ({\x0},0.70) {$\varnothing$};
\node[ghost]  at ({\x1},0.70) {$s$};
\node[simg]   at ({\x2},0.70) {$o^{w}$};
\node[simg]   at ({\x3},0.70) {$o^{p}$};
\node[ghost]  at ({\x4},0.70) {};
\fill[cAct, draw=black!55, line width=0.3pt] ({\x4},0.47) rectangle ++(0.10,0.46);
\node[anchor=west, inner sep=0pt] at ({\x4+0.15},0.70) {$a{+}s$};
\node[ghost]  at ({\x5},0.70) {$s'$};
\node[sfut]   at ({\x6},0.70) {$o'^{w}$};
\node[sfut]   at ({\x7},0.70) {$o'^{p}$};
\node[ghost]  at ({\x8},0.70) {$V$};
\draw[brace] ({\x0+0.02},0.42) -- ({\x3+0.66},0.42) node[midway, below=2pt] {clean observation};
\draw[brace] ({\x4+0.02},0.42) -- ({\x7+0.66},0.42) node[midway, below=2pt, align=center] {28 action queries + future latents};
\node[anchor=south east, inner sep=1pt] at ({\x8+0.68},0.94) {1,008 tokens};
\foreach \b in {0,...,27} {
  \pgfmathtruncatemacro{\keep}{(\b<3) || (\b>18) ? 1 : 0}
  \ifnum\keep=1
    \fill[cKeep] ({\d\b},0.47) rectangle ++(0.058,0.46);
  \else
    \draw[black!28, line width=0.3pt] ({\d\b},0.47) rectangle ++(0.058,0.46);
  \fi
}
\node[anchor=north, inner sep=1pt] at (10.45,0.46) {0.88B};
\node at (11.95,0.70) {\small\textbf{1}};
\node[inp] at (12.55,0.70) {\textbf{57\,ms}\\ \textbf{2.30\,TFLOP}\\ \textbf{2.36\,GiB}};
\end{tikzpicture}}
\caption{\textbf{From \cosmos to \smol.} Latent frames (width $\propto$ tokens), kept DiT blocks, denoising steps, and cost per 16-action chunk on an A100. \smol drops the state and value frames (dashed), replaces the action frame with 28 state-conditioned queries ($a{+}s$), and reads actions and future camera latents (its world head) from one step.}
\label{fig:overview}
\end{figure}

We present \smol (Fig.~\ref{fig:overview}).
Our premise is that much of the teacher's cost comes from an interface designed for video generation, not from what control requires.
\smol reuses the trunk's weights, token positions, and rectified-flow readout; it encodes single frames through a small adapter, replaces T5-XXL with cached BERT features~\citep{devlin2019bert} padded to preserve the trunk's attention sinks, keeps 12 of 28 blocks, and reads actions from 28 state-conditioned queries.
Most changes break the trunk zero-shot, so a distillation curriculum recovers from pruning before compaction.

We contribute (i)~\smol, which retains 97\% of its teacher's success at 11$\times$ lower latency; (ii)~a method, with a diagnostic behind each step, for deriving compact WAMs from pretrained video WAMs (Sec.~\ref{sec:method}), including three pitfalls (attention-sink padding, jointly fatal blocks, and chunks that cannot be cut short); and (iii)~findings that the compressed world head preserves motion forecasting, whereas latent-action cascades or visual-reward fine-tuning degrade control (Sec.~\ref{sec:experiments}).

%===============================================================================
\section{Related Work}
\label{sec:related}

Generalist VLAs bring pretraining to control~\citep{brohan2023rt2,kim2024openvla,openx2024,octo2024,black2025pi0,pertsch2025fast}, often with diffusion~\citep{chi2023diffusionpolicy} or chunked~\citep{zhao2023act} action heads; OpenVLA-OFT~\citep{kim2025openvlaoft}, TinyVLA~\citep{wen2025tinyvla}, SmolVLA~\citep{shukor2025smolvla}, and TurboVLA~\citep{xie2026turbovla} speed them up; \smol borrows TurboVLA's principles but retains a world model.
Latent world models learn dynamics in feature space~\citep{zhou2025dinowm,assran2025vjepa2,zheng2025flare,maes2026leworldmodel,gao2026fastleworldmodel}, and GR-1~\citep{wu2024gr1}, PAD~\citep{guo2024pad}, and WorldVLA~\citep{cen2025worldvla} generate future images and actions jointly.
Recent WAMs build on video models~\citep{kim2026cosmospolicy,ye2026dreamzero} and add action evaluation~\citep{zhou2026tau0wm}, modalities~\citep{nvidia2026cosmos3}, object state~\citep{liu2026oawam}, geometry~\citep{lopetegui2026sawam}, or latent actions~\citep{chen2026abotm05}; Fast-WAM~\citep{yuan2026fastwam} drops future prediction at test time and Light-WAM~\citep{li2026lightwam} decodes actions from a smaller video backbone, whereas \smol reads future latents from its action pass.

%===============================================================================

\section{SmolWAM}
\label{sec:method}

\subsection{Teacher: Cosmos Policy}

On LIBERO, the video diffusion transformer (DiT)~\citep{peebles2023dit} of \cosmos (28 blocks, 1.96B parameters) denoises nine latent frames of $14{\times}14{=}196$ tokens (Fig.~\ref{fig:overview}): a blank frame, the robot state $s$, the wrist and third-person camera latents $o^{w}$ and $o^{p}$, a chunk $a$ of $H{=}16$ actions, and, at time $t{+}H$, the state $s'$, the camera latents $o'^{w}$ and $o'^{p}$, and a value $V$.
Images are encoded by the Wan2.1 VAE~\citep{wan2025} from 33-frame clips, and the instruction enters through cross-attention to 512 T5-XXL embeddings.
The model is a rectified flow~\citep{liu2023rectifiedflow,lipman2023flowmatching}: for $x_\tau=(1-\tau)\,x_0+\tau\,\epsilon$ it predicts the velocity $v_\theta(x_\tau,\tau)\approx\epsilon-x_0$, which gives the closed-form clean estimate $\hat{x}_0 = x_\tau - \tau\, v_\theta(x_\tau,\tau)$.
Generated frames start at $\tau_{\max}\approx0.988$ ($\sigma_{\max}{=}80$); the released pipeline takes five steps.

\subsection{Deriving SmolWAM from the teacher}
\label{sec:derivation}

\textbf{Single-frame latents.}
\smol encodes each camera image once, as a single frame, with the Wan2.1 encoder (54M parameters).
Such latents raise the teacher's action error (normalized L1 of its one-step chunk to the demonstration) from 0.0089 to 0.038; a 0.5M-parameter adapter maps them to clip-mode latents and restores it to 0.0092.

\textbf{Cached text that keeps the attention sinks.}
We replace T5-XXL with cached BERT-base~\citep{devlin2019bert} features of the instruction, projected to the context width.
The trunk's cross-attention has bias-free key and value projections, so T5's ${\sim}500$ zero-padded positions contribute zero logits and values and act as an attention sink~\citep{xiao2024streamingllm}: truncating the context to 32 or 128 positions drops the teacher to 0/5 closed-loop successes, and restoring the removed softmax mass analytically returns it to 5/5, matching the padded attention to $6{\times}10^{-8}$.
\smol thus zero-pads the BERT features to 512 positions; six zero-initialized vision-to-language layers from TurboVLA~\citep{xie2026turbovla} let the text attend to the images.

\textbf{Boundary-preserving pruning.}
Block influence $1-\cos(h_{\text{in}},h_{\text{out}})$~\citep{men2025shortgpt} is highest for blocks 0, 1, and 25--27, while blocks 3--22 are individually close to the identity.
Zero-shot screens (5 episodes each) show that removals do not compose: skipping block 1, 26, or 27 fails every episode, skipping block 7, 14, or 20 alone fails none, and skipping four near-identity blocks (12--15) together fails all.
\smol keeps blocks 0--2 and 19--27 (0.85B parameters).

\textbf{State-conditioned action queries.}
A zero-initialized MLP adds the robot state to 28 action queries that replace the 196-token action frame at its original positions.
Beyond these queries we keep only the blank and image frames (1,008 tokens).
Zero-shot, dropping the future-state and value frames barely changes the action error (0.009 to ${\le}0.019$), but dropping the blank (0.30) or state (0.058) frame or compacting the vector frames to 28 tokens (0.21) does, so the layout is learned.

\textbf{One-pass readout.}
\smol evaluates its trunk once at $\tau_{\max}$ with fixed noise and reads the action chunk and future latents from $\hat{x}_0$, so its actions never condition on a denoised future; as Fast-WAM~\citep{yuan2026fastwam} also finds, acting does not need one, and the teacher solves all 60 screening episodes in one step (Table~\ref{tab:main}).
The future latents form its world head, supervised in training and never decoded.

\subsection{Distillation with a layout curriculum}
\label{sec:training}

We train on 68,830 observations sampled every fourth step from the 1,700 successful LIBERO demonstrations released with \cosmos, using the teacher's one-step predictions of the action chunk and future latents, $a^{\mathcal{T}}$ and $z^{\mathcal{T}}$, cached at fixed noise (66 min on one A100).
As in pruning and distillation of language models~\citep{muralidharan2024minitron,hinton2015distilling}, the loss $\|\hat a - a\|_1 + \|\hat a - a^{\mathcal{T}}\|_1 + \|\hat z - z\|_1 + \|\hat z - z^{\mathcal{T}}\|_1$ matches the student's one-step estimates to both the demonstrations and the teacher, plus 0.1-weighted terms for the state and value frames while they exist.
Pruning and compaction are separate distribution shifts, so the student, initialized from the kept teacher blocks, trains in the teacher's 1,764-token layout for the first 1,000 of 5,000 steps and in the compact layout afterward (AdamW; learning rate $2{\times}10^{-5}$ for inherited and $10^{-4}$ for new weights; batch 32; bf16) in 3.6\,h on one A100 (App.~\ref{app:training}).

%===============================================================================
\section{Experiments}
\label{sec:experiments}

We ask \textbf{(Q1)} how much of the teacher's success \smol retains, \textbf{(Q2)} where its savings come from, \textbf{(Q3)} whether it remains a world model, and \textbf{(Q4)} whether richer WAM components help.

\textbf{Setup.}
LIBERO~\citep{liu2023libero} has four suites of ten tasks; we use the teacher's observation pipeline (wrist and third-person images at 224\,px, proprioception, 16-action chunks executed open loop), seed 195, and 10 trials per task (400 episodes).
A 60-episode screening set (3 tasks per suite, 5 trials each) includes the tasks an earlier pruned student failed most, so it scores below the full protocol.
Latency is the wall-clock time from observation to action chunk on one A100-SXM4-40GB (batch 1, bf16), including preprocessing and VAE encoding, with text features cached for every model; FLOPs are counted with PyTorch's \texttt{FlopCounterMode}, attention included.

\begin{table}[t]
\centering
\caption{\textbf{LIBERO success and inference cost.} Screen: 60 episodes; suites and Full: 400. Cost per 16-action chunk on one A100 (batch 1, bf16); Params excludes the VAE and text encoder; TFLOP counts network and VAE encoder; GiB: peak memory. $^\ast$Unmodified weights, single-frame latents via the 0.5M-parameter adapter. $^\dagger$Suites and Full from an earlier evaluation pipeline, on an L4 (\cosmos reports 98.5\%~\citep{kim2026cosmospolicy}). $^\ddagger$Shared GPU. $^\S$Simulated: INT8 weights, bf16 kernels.}
\label{tab:main}
\small
\setlength{\tabcolsep}{2.85pt}
\begin{tabular}{@{}lrrrrrrrrrr@{}}
\toprule
& \multicolumn{4}{c}{Inference cost} & \multicolumn{6}{c}{Success rate (\%)} \\
\cmidrule(lr){2-5}\cmidrule(l){6-11}
Model & Params & TFLOP & ms & GiB & Screen & Spatial & Object & Goal & Long & Full \\
\midrule
\cosmos (5 steps) & 1.96B & 42.6 & 624 & 7.55 & 96.7 & 97.0 & 100.0 & 99.0 & 97.0 & 98.2$^\dagger$ \\
\cosmos (1 step, 1 frame)$^\ast$ & 1.96B & 7.52 & 126 & 4.39 & 100.0 & -- & -- & -- & -- & -- \\
Pruned + LoRA distillation & 1.40B & -- & -- & -- & -- & 83.0 & 100.0 & 82.0 & 85.0 & 87.5$^\dagger$ \\
\midrule
\textbf{\smol} & \textbf{0.88B} & \textbf{2.30} & \textbf{57} & \textbf{2.36} & 93.3 & 99.0 & 100.0 & 94.0 & 89.0 & \textbf{95.5} \\
\quad + 7.5k training steps & 0.88B & 2.30 & 60 & 2.36 & 93.3 & 94.0 & 100.0 & 94.0 & 89.0 & 94.2 \\
\quad INT8 weights$^\S$ & 0.88B & 2.30 & 65$^\ddagger$ & 2.36 & 91.7 & -- & -- & -- & -- & -- \\
\quad 8 blocks & 0.60B & 1.76 & 134$^\ddagger$ & 1.84 & 83.3 & -- & -- & -- & -- & -- \\
\smolla (4 steps) & 0.92B & 4.28 & 239 & 2.43 & 53.3 & -- & -- & -- & -- & -- \\
\bottomrule
\end{tabular}
\end{table}

\textbf{Q1: \smol retains most of the teacher's success.}
\smol succeeds in 382/400 episodes (95.5\%; 95\% Wilson interval 93.0--97.1) against the teacher's 393/400 (98.2\%; 96.4--99.1); App.~\ref{app:rollouts} shows both from the same initial states.
We detect no difference on Spatial and Object; the deficit lies in the multi-step Goal (94 vs.\ 99) and Long (89 vs.\ 97) suites, where failed rollouts usually complete the first subgoal and stall on the second.
An earlier student that removed only 8 blocks and recovered with LoRA~\citep{hu2022lora} distillation (19.7M trainable parameters, under an hour on one L4, no demonstrations) on 414 teacher-rollout observations reached 87.5\%; \smol differs in distillation coverage, full fine-tuning, and the learned layout, effects we have not separated.
Longer training does not help, and 8 blocks fall to 83.3\% on the screen (Table~\ref{tab:main}).

\textbf{Q2: Where the 11$\times$ comes from.}
Running the teacher's unmodified weights for one step on single-frame inputs cuts latency 5$\times$ and compute 5.7$\times$, to 126\,ms and 7.5\,TFLOP, with no loss on the screening set; with no adapter and the text truncated under the exact sink correction, the released model runs 8$\times$ faster on an L4 (1.83 to 0.23\,s; no failures in an 11-episode screen; App.~\ref{app:ladder}).
\smol's trunk redesign adds 2.2$\times$ in latency, 3.3$\times$ in compute, and 1.9$\times$ in memory; of its 57\,ms, the transformer takes 37, the VAE encode 13, and preprocessing 7 (success against cost in App.~\ref{app:pareto}).
At 57\,ms per chunk it issues 17.7 chunks (282 actions) per second; because chunks execute open loop, the chunk rate bounds how often the policy reacts to new observations.
Cutting chunks short does not help: re-querying the teacher after 8 of its 16 actions solves no screening episode at one or five denoising steps, so faster reaction needs training-time changes~\citep{black2025rtc}.

\begin{wrapfigure}[9]{r}{0.36\linewidth}
\vspace{-12pt}
\centering
{\scriptsize\makebox[0.333\linewidth]{current}\makebox[0.333\linewidth]{predicted}\makebox[0.333\linewidth]{real}}\\[1pt]
\includegraphics[width=\linewidth]{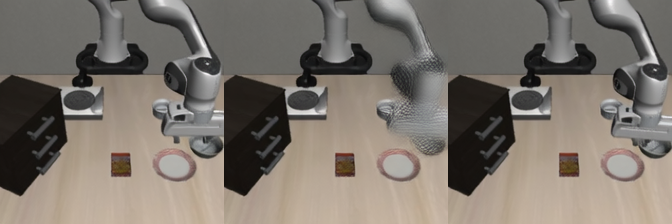}
\caption{\textbf{World-head forecast}, decoded only for display.}
\label{fig:dreams}
\end{wrapfigure}
\textbf{Q3: \smol remains a world model.}
On 96 held-out contexts, \smol's forecast of the view 16 steps ahead lowers latent L1 by 24\% relative to copying the current frame (0.101 vs.\ 0.133), and lowers the end-point error between RAFT~\citep{teed2020raft} optical flow to the predicted and to the real future from 1.63 to 1.43 pixels.
Decoded forecasts (Fig.~\ref{fig:dreams}) keep the scene and place the arm near its future pose but blur its appearance, and LPIPS~\citep{zhang2018lpips} stays at the copy-frame level (0.128 vs.\ 0.115): the world head captures motion rather than texture.

\textbf{Q4: Richer WAM components hurt at this scale.}
Porting ABot-M0.5~\citep{chen2026abotm05} onto \smol (latent-action tokens, mixture-of-transformer towers, four-step flow matching, and dream forcing) yields \smolla, which succeeds in 53.3\% of screening episodes at 239\,ms, and reward fine-tuning of its world head collapses success to 5.0--16.7\% while worsening every image metric (App.~\ref{app:richer}).

%===============================================================================
\section{Limitations and Conclusion}
\label{sec:limitations}

\textbf{Limitations.}
We evaluate SmolWAM only on LIBERO; future work should extend evaluation to additional robotics benchmarks such as RoboCasa and, ultimately, real-robot settings. We also observe that reducing transformer depth disproportionately affects long-horizon tasks, with the largest degradation on LIBERO-Long. Developing pruning strategies that preserve long-horizon control performance therefore remains an important open problem.

\textbf{Conclusion.}
A pretrained video WAM tolerates large interface and depth changes when each is distilled rather than applied zero-shot, but not auxiliary objectives that train its shared trunk without an action anchor; \smol keeps its world prediction and 95.5\% LIBERO success at a fraction of the teacher's cost.

\bibliographystyle{plainnat}
\bibliography{refs}

%===============================================================================
\appendix
\raggedbottom  % the appendix's figures stay in place ([H]); end short pages with space, not stretched gaps
\section{Additional results}
\label{app}

\subsection{Success against cost}
\label{app:pareto}

\begin{figure}[H]
\centering
\includegraphics[width=\textwidth]{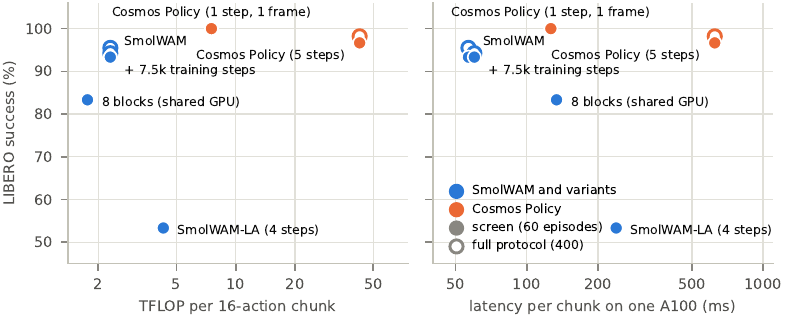}
\caption{\textbf{LIBERO success against cost} for every row of Table~\ref{tab:main} with an A100 cost. Filled markers are the 60-episode screen and hollow markers the 400-episode protocol (the teacher's as in Table~\ref{tab:main}); the 8-block latency was measured on a shared GPU.}
\label{fig:pareto}
\end{figure}

\smol and its 7.5k-step continuation sit in the cheap, successful corner. The 8-block variant saves a further 0.5\,TFLOP but loses 10 points on the screen, and \smolla is both slower and less successful than \smol. Per-channel INT8 quantization of the 172 largest linear layers shrinks the stored network from 1.64 to 0.85\,GiB and costs one screening episode; simulated with bf16 kernels, it does not yet reduce latency.

\subsection{Training curves}
\label{app:training}

\begin{figure}[H]
\centering
\includegraphics[width=\textwidth]{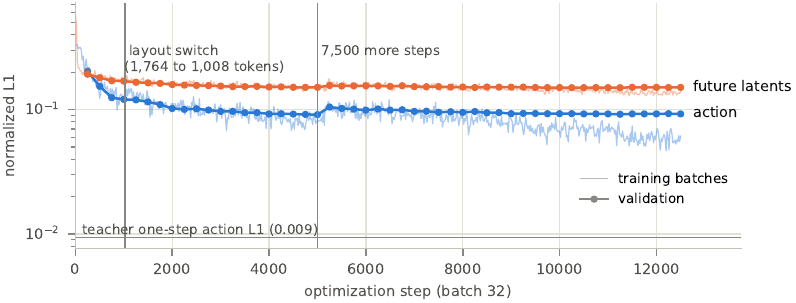}
\caption{\textbf{Distillation of \smol} (batch 32): normalized L1 of the one-step action chunk and future latents against the demonstrations, on training batches and held-out validation rows. Vertical lines mark the switch from the teacher's 1,764-token layout to the compact one, and the start of 7,500 further steps.}
\label{fig:training}
\end{figure}

Validation action L1 falls fastest in the teacher's layout, keeps falling after the switch to the compact layout, and settles near 0.09, ten times the teacher's own one-step error, while \smol still succeeds in 95.5\% of episodes: offline L1 is a poor proxy for closed-loop success. The 7,500 further steps lower the training loss but neither the validation loss nor success (Table~\ref{tab:main}).

\subsection{The zero-training ladder on an L4}
\label{app:ladder}

\begin{table}[H]
\caption{\textbf{Zero-training speedups on the released weights} (L4, uncontended single-query benchmark). Deviation: mean absolute difference from the five-step model's actions (raw units) on the benchmark's single observation; over a 161-observation offline set, one step deviates by 0.003. Closed-loop scores are single-trial screens, so they screen rather than resolve small differences.}
\label{tab:ladder}
\centering
\small
\begin{tabular}{lccl}
\toprule
Configuration (cumulative) & s/chunk & deviation & closed loop \\
\midrule
\cosmos 2B, 5 denoising steps (as released) & 1.83 & -- & 5/5, 3/3 Spatial \\
1 denoising step & 1.22 & 0.005 & 5/5 \\
+ no future-image decode & 0.54 & 0.005 & 5/5 \\
+ single-frame VAE encode + exact text sink & \textbf{0.23} & 0.044 & 5/5, 3/3 Spatial, 3/3 Goal \\
\bottomrule
\end{tabular}
\end{table}

Each change is applied to the released model with no training. At one step, a query on the L4 spends 0.2\,s in the transformer, 0.35\,s encoding 33-frame clips, and 0.68\,s decoding future images that the controller never reads. The released evaluation pipeline decodes them; the teacher pipeline of Table~\ref{tab:main} does not.

\subsection{Richer WAM components}
\label{app:richer}

\smolla adds 39M parameters to \smol: tokens from a frozen 32-dimensional latent-action model behind an asymmetric attention mask, mixture-of-transformer towers for the latent-action, arm, and gripper streams, four-step conditional flow matching, and dream forcing, which trains the action head on predicted latents. Its one-step action estimate matches \smol's offline (validation L1 0.089 vs.\ 0.091), but four-step sampling raises the error to 0.155, and sampling with one or two steps recovers only 63.3\% and 70.0\% of screening episodes; its latent-action predictions are worse than predicting the mean code (MSE 1.15, variance 0.64), and its world head is no better (latent L1 0.102). Reward-weighted fine-tuning of the world head under five rewards built from optical flow, depth~\citep{yang2024depthanythingv2}, and LPIPS worsened every image metric (LPIPS from 0.130 to 0.167--0.177) and collapsed success to 5.0--16.7\% within 300 steps, likely because the objective has no action term while the shared trunk trains.

\subsection{Single-flag ablations}
\label{app:ablations}

\begin{figure}[H]
\centering
\includegraphics[width=\textwidth]{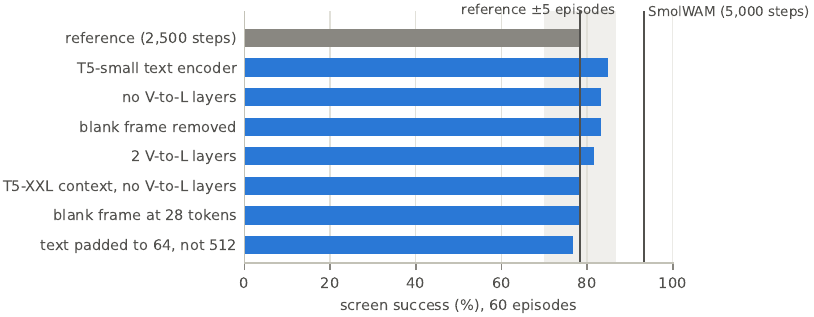}
\caption{\textbf{Single-flag ablations} of the \smol recipe, each trained for 2,500 steps (half the main run) and scored on the 60-episode screen. The shaded band spans five episodes either side of the reference.}
\label{fig:ablations}
\end{figure}

Every change lands within five episodes of the reference (78.3\%), so at this budget the screen cannot separate these choices; the full-budget model scores 93.3\% on the same screen. Six planned ablations, among them removing the world head, the distillation targets, and the layout curriculum, were not run.

\subsection{Rollouts}
\label{app:rollouts}

\begin{figure}[H]
\centering
\includegraphics[width=\textwidth]{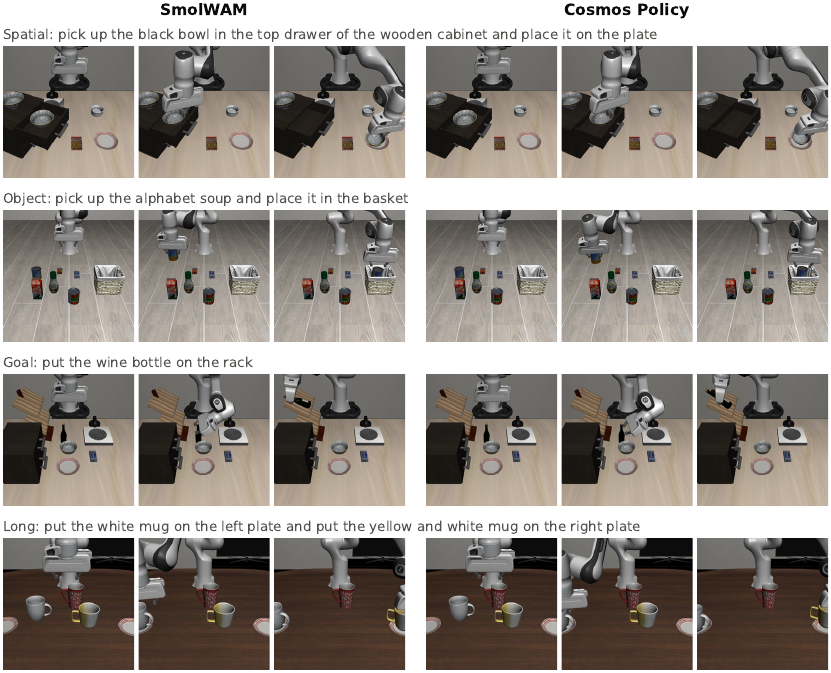}
\caption{\textbf{Rollouts} of \smol (left) and \cosmos (right) from the same initial state, one task per suite: first, middle, and last frames. Both succeed on all four, in 131, 147, 154, and 234 steps (\smol) and 143, 147, 138, and 247 steps (\cosmos). Recorded on an Apple M2 Pro.}
\label{fig:rollouts}
\end{figure}

\end{document}